\RequirePackage{fix-cm}
\documentclass[smallcondensed]{svjour3}       
\smartqed  
\usepackage{graphicx}
\usepackage{graphbox}
\usepackage[ruled,vlined,linesnumbered,titlenumbered]{algorithm2e}
\usepackage{multirow}
\usepackage{amssymb,stmaryrd}
\usepackage{xparse,tikz}
\usepackage{url}
\usepackage{booktabs}
\usepackage{hhline}
\usepackage{enumitem}
\usepackage{array}
\usepackage{makecell}
\usepackage{balance}
\usepackage{subfigure}
\usepackage{amsmath}
\usepackage{textcomp}

\newcommand\algname{CURIE}

\DeclareMathOperator*{\argmax}{arg\,max}

\begin{document}

\title{\algname: A Cellular Automaton for Concept Drift Detection
}


\dedication{Dedicated to Tom Fawcett and J. H. Conway, who passed away in 2020, for their noted contributions to the field of cellular automata and machine learning, and for inspiring this research.}

\author{Jesus L. Lobo \and
        Javier Del Ser \and
        Eneko Osaba \and
        Albert Bifet \and
        Francisco Herrera
}


\institute{Jesus L. Lobo \at
              TECNALIA, Basque Research and Technology Alliance (BRTA), \\
              \email{jesus.lopez@tecnalia.com}           
			\and 
			Javier Del Ser \at 
			University of the Basque Country UPV/EHU \& Basque Center for Applied Mathematics (BCAM)  \\
			\email{javier.delser@ehu.eus, jdelser@bcamath.org}
			\and 
			Eneko Osaba \at 
			TECNALIA, Basque Research and Technology Alliance (BRTA), \\
			\email{eneko.osaba@tecnalia.com}
           \and
		   Albert Bifet \at
		   ParisTech \& The University of Waikato \\
		   \email{albert.bifet@telecom-paristech.fr, abifet@waikato.ac.nz}           
           \and
		   Francisco Herrera \at
		   Andalusian Research Institute in Data Science and Computational Intelligence (DaSCI), \\ 
		   University of Granada \\
		   \email{herrera@decsai.ugr.es}           			  
}

\date{Received: 09-19-2020 / Accepted: date}

\maketitle

\begin{abstract}
Data stream mining extracts information from large quantities of data flowing fast and continuously (data streams). They are usually affected by changes in the data distribution, giving rise to a phenomenon referred to as \textit{concept drift}. Thus, learning models must detect and adapt to such changes, so as to exhibit a good predictive performance after a drift has occurred. In this regard, the development of effective drift detection algorithms becomes a key factor in data stream mining. In this work we propose $\algname$, a drift detector relying on cellular automata. Specifically, in $\algname$ the distribution of the data stream is represented in the grid of a cellular automata, whose neighborhood rule can then be utilized to detect possible distribution changes over the stream. Computer simulations are presented and discussed to show that $\algname$, when hybridized with other base learners, renders a competitive behavior in terms of detection metrics and classification accuracy. $\algname$ is compared with well-established drift detectors over synthetic datasets with varying drift characteristics.
\end{abstract}

\section{Introduction}\label{intro}

Data Stream Mining (DSM) techniques are focused on extracting patterns from continuous (potentially infinite) and fast data. A data stream is the basis of machine learning techniques for this particular kind of data, which is composed of an ordered sequence of instances that arrive one by one or in batches. Depending on the constraints imposed by the application scenario at hand, such instances can be read only once or at most a reduced number of times, using limited computing and memory resources. These constraints require an incremental learning (or one-pass learning) procedure where past data cannot be stored for batch training in future time steps. Due to these challenging conditions under which learning must be done, DSM has acquired a notable relevance in recent years, mostly propelled by the advent of Big Data technologies and data-intensive practical use cases \cite{MOA-Book-2018}. 

In this context, data streams are often generated by non-stationary phenomena, which may provoke a change in the distribution of the data instances (and/or their annotation). This phenomenon is often referred to as \textit{concept drift} \cite{webb2016characterizing}. These changes cause that predictive models trained over data streams become eventually obsolete, not adapting suitably to the new distribution (concept). The complexity of overcoming this issue, and its prevalence over many real scenarios, make \textit{concept drift} detection and adaptation acknowledged challenges in DSM \cite{lu2018learning}. Examples of data stream sources undergoing \textit{concept drift} include computer network traffic, wireless sensor data, phone conversations, social media, marketing data, ATM transactions, web searches, and electricity consumption traces, among others \cite{vzliobaite2016overview}. Recently, several emerging paradigms such as the so-called Smart Dust \cite{ilyas2018smart}, Utility Fog \cite{dastjerdi2016fog}, Microelectromechanical Systems (MEMS or ``motes") \cite{judy2001microelectromechanical}, or Swarm Intelligence and Robotics \cite{del2019bio}, are in need for efficient and scalable solutions in real-time scenarios. Here \textit{concept drift} may be present, and thus making drift detection a necessity. 

This complexity in the \textit{concept drift} phenomenon manifests when researchers try to characterize it \cite{webb2016characterizing}. Indeed, there are many different types of concept drifts, characterized by e.g. the speed or severity of change. Consequently, drift detection is a key factor for those active strategies that require triggering mechanisms for drift adaptation \cite{hu2019no}. A drift detector estimates the time instants at which changes occur over the stream, so that when a change is detected, an adaptation mechanism is applied to the base learner so as to avoid a degradation of its predictive performance. The design of a \textit{concept drift} detector with high performance is not trivial, yet it is crucial to achieve more reliable DSM models. In fact, a general-purpose strategy for \textit{concept drift} detection, handling and recovery still remains as an open research avenue, as foretold by the fulfillment of the No Free Lunch theorem in this field \cite{hu2019no}. This difficulty to achieve a universal best approach becomes evident in the most recent comparative among drift detectors made in \cite{barros2018large}. Analyzing its mean rank of methods, we observe how there is no a method with the best metrics, or even showing the best performance in the majority of them. In this regard, the design objective is to develop techniques that detect all existing drifts in the stream with low latency and as few false alarms and missed detections as possible. Thus, the most suitable drift detector depends on the characteristics of the DSM problem under study, giving more emphasis to some metrics than others. Regarding the detection metrics, we usually tend to put in value those drift detectors that are able to show a good classification performance while minimizing the distance of the true positive detections. 

Cellular automata (CA), as low-bias and robust-to-noise pattern recognition methods with competitive classification performance, meet the requirements imposed by the aforementioned paradigms mainly due to their simplicity and parallel nature. In this work we present a Cellular aUtomaton for concept dRIft dEtection ($\algname$), capable of competitively identifying drifts over data streams. The proposed approach is based on CA, which became popular when Conway's Game of Life appeared in $1970$, and thereafter attracted attention when Stephen Wolfram published his CA study in $2002$ \cite{wolfram2002new}. Although CA are not very popular in the data mining field, Fawcett showed in \cite{fawcett2008data} that they can become simple, low-bias methods. $\algname$, as any other CA-based technique, is computationally complete (able to perform any computation which can be done by digital computers) and can model complex systems from simple structures, which puts it in value to be considered in the DSM field. Moreover, $\algname$ is tractable and interpretable \cite{lobo2020lunar}, both ingredients that have lately attracted attention under the eXplainable Artificial Intelligence (XAI) paradigm \cite{arrieta2020explainable}.  Next, we summarize the main contributions of $\algname$ to the drift detection field:
\begin{itemize}[leftmargin=*]
	\item It is capable of competitively detecting abrupt and gradual concept drifts. 
	\item It does not required the output (class prediction) of the base learner. Instead, it extracts the required information for drift detection from its internal structure, looking at the changes occurring in the neighborhood of cells.
	\item It is interpretable, due to the fact that its cellular structure is a direct representation of the feature space and the labels to be predicted.
	\item It can be combined with any base learner.
\end{itemize} 
Besides, $\algname$ offers other additional advantage in DSM:
\begin{itemize}[leftmargin=*]
	\item It is also able to act as an incremental learner and adapt to the change \cite{lobo2020lunar}, going one step further by embodying an \textit{all-in-one} approach (learner and detector).	
\end{itemize} 
The rest of the manuscript is organized as follows: first, we provide the background of the field in Sect. \ref{related_work}. Next, we introduce the fundamentals of CA and their role in DSM in Sect. \ref{cela}. Sect. \ref{proposed} exposes the details of our proposed drift detector $\algname$, whereas Sections \ref{expems} and \ref{res} elaborate on experimental setup and analyze results with synthetic and real-world data stream respectively. Finally, Sect. \ref{concs} concludes the manuscript with an outlook towards future research derived from this work.

\section{Related Work}\label{related_work}

We now delve into the background literature related to the main topics of this work: drift detection (Subsection \ref{sec:dd}) and cellular automata for machine learning (Subsection \ref{sec:caml}).

\subsection{Drift Detection} \label{sec:dd}

DSM has attracted much attention from the machine learning community \cite{gomes2019machine}. Researchers are now on the verge of moving out DSM methods from laboratory environments to real scenarios and applications, similarly to what occurred with traditional machine learning methods in the past. Most efforts in DSM have been focused on supervised learning \cite{MOA-Book-2018} (mainly on classification), addressing the \textit{concept drift} problem \cite{webb2016characterizing}. Generally, these efforts have been invested in the development of new methods and algorithms that maintain an accurate decision model with the capability of learning incrementally from data streams while forgetting concepts \cite{losing2018incremental,wares2019data}. 

For this purpose, drift detection and adaptation mechanisms are needed \cite{lu2018learning}. In contrast to passive (blind) approaches where the model is continuously updated every time new data instances are received (i.e., drift detection is not required), active strategies (where the model gets updated only when a drift is detected) are in need for effective drift detection mechanisms. Most active approaches usually utilize a specific classifier (base learner) and analyze its classification performance (e.g. accuracy or error rate) to indicate whether a drift has occurred or not. Then, the base learner is trained on the current instance within an incremental learning process repeated for each incoming instance of the data stream. Despite the most used input for the drift detectors are the accuracy or error rate, we can find other detectors that use other inputs such as diversity \cite{minku2011ddd} or structural changes stemming from the model itself \cite{lobo2018drift}.

There is a large number of drift detectors in the literature, many of them compared in \cite{gonccalves2014comparative}. As previously mentioned, the conclusion of these and other works is that there is no a general-purpose strategy for \textit{concept drift}. The selection of a good strategy depends on the type of drift and particularities of each data streaming scenario. Other more recent \textit{concept drift} detection mechanisms have been presented and well described in \cite{barros2018large}. 

\subsection{Cellular Automata for Pattern Recognition} \label{sec:caml}

CA are not popular in the pattern recognition community, but even so we can find recent studies and applications. In \cite{collados2019distributed}, authors propose CA to simulate potential future impacts of climate change on snow covered areas, whereas in \cite{gounaridis2019random} an approach to explore future land use/cover change under different socio-economic realities and scales is presented. Scheduling is another field where CA has been profusely in use \cite{carvalho2019improving}. Another recent CA approach for classification is \cite{uzun2018solution}. CA have been also used with convolutional neural networks \cite{gilpin2019cellular} and reservoir computing \cite{nichele2017deep}.

Regarding DSM or \textit{concept drift} detection fields, the presence of CA-based proposals is even scarcer. Although a series of contributions unveiled solid foundations for CA to be used for pattern recognition \cite{jen1986invariant,raghavan1993cellular,chaudhuri1997additive}, it was not until $2008$ \cite{fawcett2008data} (departing from the seminal work in \cite{ultsch2002data}) when CA was presented as a simple but competitive method for parallelism, with a low-bias, effective and competitive in terms of classification performance, and robust to noise. Regarding DSM and \textit{concept drift} detection, timid efforts have been reported so far in \cite{hashemi2007better} and \cite{pourkashani2008cellular}, which must be considered as early attempts to deal with noise rather than with incremental learning and drift detection. They used a CA-based approach as a real-time instance selector to avoid noisy instances, while the classification task was performed in batch learning mode by non-CA-based learning algorithms. Thus, CA is proposed as a mere complement to select instances, and not as an incremental learner. Besides, their detection approach is simply based on the local class disagreements between neighboring cells, without considering relevant aspects such as the grid size, the radius of the neighborhood, or the moment of the disagreement, among other factors. Above all, they do not provide any evidence on how their solution learns incrementally, nor details on the real operation of the drift detection approach. Finally, in terms of drift detection evaluation, their approach is not compared to known detectors using reputed base learners and standard detection metrics.

More recently, the authors of \cite{lobo2020lunar} transform a cellular automaton into a real incremental learner with drift adaptation capabilities. In this work, we go one step further by proposing $\algname$, a cellular automaton featuring a set of novel ingredients that endow it with abilities for drift detection in DSM. As we will present in detail, $\algname$ is an interpretable CA-based drift detector, able to detect abrupt and gradual drifts, and providing very competitive classification performances and detection metrics.

\section{Cellular Automata}\label{cela}

\subsection{Foundations}
Von Neumann described CA as discrete dynamical systems with a capacity of universal computability \cite{von1966theory}. Their simple local interaction and computation of cells result in a huge complex behavior when these cells act together, being able to describe complex systems in several scientific disciplines. 

\begin{figure}[h]
	\centering
	\begin{tabular}{cc}
		\includegraphics[width=0.25\linewidth]{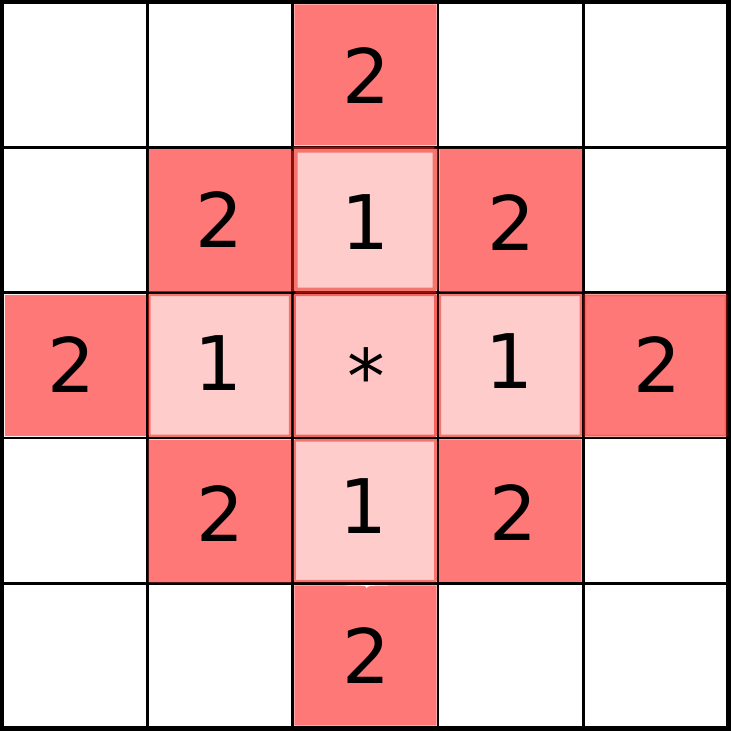} &
		\includegraphics[width=0.4\linewidth]{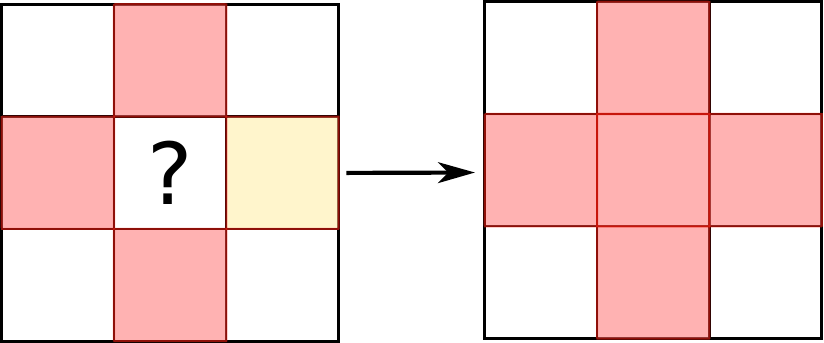} \\
		\small (a) & \small (b)\\
	\end{tabular}
	\caption{Neighborhood of CA in data mining: (a) a von Neumann's neighborhood with radius $r=1$ and $r=2$ using the Manhattan distance; (b) the center cell inspects its \emph{von Neumann's} neighborhood ($r=1$) and applies the majority voting rule in a one-step update.}
	\label{vn_neigh}
\end{figure}

Following the notation of Kari in \cite{kari2005theory}, a cellular automaton can be formally defined as: \smash{$A \doteq (d,\mathcal{S},f_{\boxplus},f_{\circlearrowright})$}, with $d$ denoting the dimension, $\mathcal{S}$ a group of discrete states, \smash{$f_{\boxplus}(\cdot)$} a function that receives as input the coordinates of the cell and returns the neighbors of the cell to be utilized by the update rule, and \smash{$f_{\circlearrowright}(\cdot)$} a function that updates the state of the cell at hand as per the states of its neighboring cells. Hence, for a radius $r=1$ \emph{von Neumann's} neighborhood defined over a $d=2$-dimensional grid, the set of neighboring cells and state of the cell with coordinates $\mathbf{c}=[i,j]$ are given by:
\begin{align*}
&f_{\boxplus}([i,j])=\{[i,j+1],[i-1,j],[i,j-1],[i+1,j]\}, \\
&S(\mathbf{c})=S([i,j])\nonumber\\
&=f_{\circlearrowright}(S([i,j+1]),S([i-1,j]),S([i,j-1]),S([i+1,j])),
\end{align*}
i.e., the vector of states $S([i,j])$ of the $[i,j]$ cell within the grid is updated according to the local rule \smash{$f_{\circlearrowright}(\cdot)$} when applied over its neighbors given by $f_{\boxplus}([i,j])$ (Figure \ref{vn_neigh}). For a $d$-dimensional space, a \emph{von Neumann's} neighborhood contains $2d$ cells. 

A cellular automaton should present these three properties: i) \textit{parallelism} or \textit{synchronicity} (all of the updates to the cells compounding the grid are performed at the same time); ii) \textit{locality} (when a cell $[i,j]$ is updated, its state $S[i,j]$ is based on the previous state of the cell and those of its nearest neighbors); and iii) \textit{homogeneity} or \textit{properties-uniformity} (the same update rule \smash{$f_{\circlearrowright}(\cdot)$} is applied to each cell).

\subsection{Cellular Automata for Data Stream Mining}\label{CA_dat_min}
A DSM process that may evolve over time can be defined as follows: given a time period $[0,t]$, the historical set of instances can be denoted as $\mathbf{D}_{0,t}={\mathbf{d}_{0},\ldots, \mathbf{d}_{t}}$, where $\mathbf{d}_{i}=(\mathbf{X}_{i},y_{i})$ is an instance, $\mathbf{X}_{i}$ is the vector of features, and $y_{i}$ its label. Assuming that $\mathbf{D}_{0,t}$ follows a certain joint probability distribution $P_{t}(\mathbf{X},y)$. As it has already been mentioned, data streams usually suffer from \textit{concept drift}, which may change their data distribution, provoking that predictive models trained over them become obsolete. Thus, \textit{concept drift} happens at timestamp $t+1$ when $P_{t}(\mathbf{X},y) \neq P_{t+1}(\mathbf{X},y)$, i.e. as a change of the joint probability distribution of $\mathbf{X}$ and $y$ at time $t$. 

In addition to the presence of \textit{concept drift}, DSM also imposes by itself its own restrictions, which calls for a redefinition of the previous CA for data mining. Algorithms learning from data streams must operate under a set of restrictions \cite{domingos2003general}:
\begin{itemize}[leftmargin=*]
	\item Each instance of the data stream must be processed only once.
	\item The time to process each instance must be low.
	\item Only a few data stream instances can be stored (limited memory).
	\item The trained model must provide a prediction at any time.
	\item The distribution of the data stream may evolve over time.
\end{itemize}

\begin{figure}[h]
	\centering
	\begin{tabular}{cc}
		\includegraphics[width=0.3\linewidth]{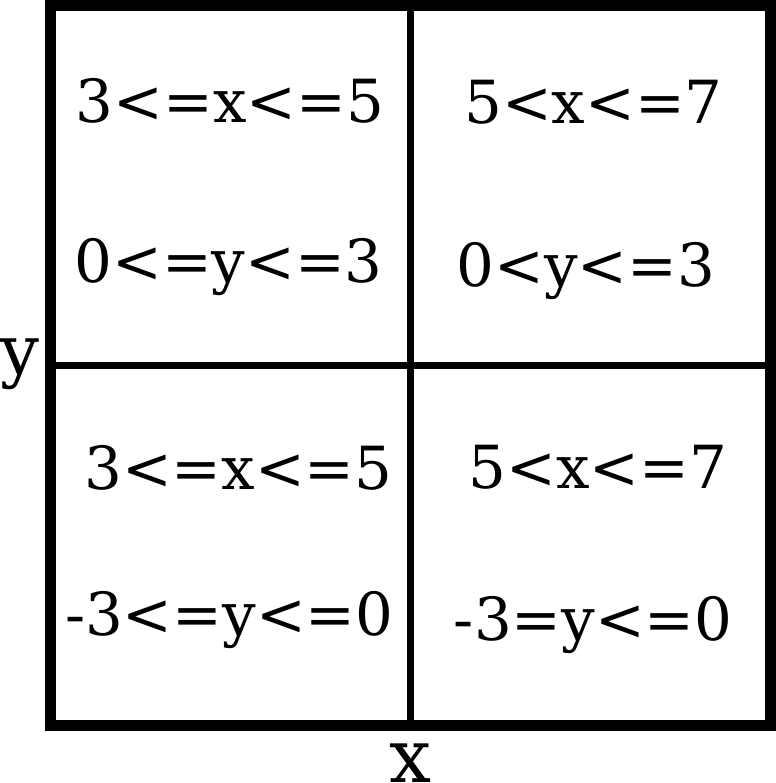} &
		\includegraphics[width=0.55\linewidth]{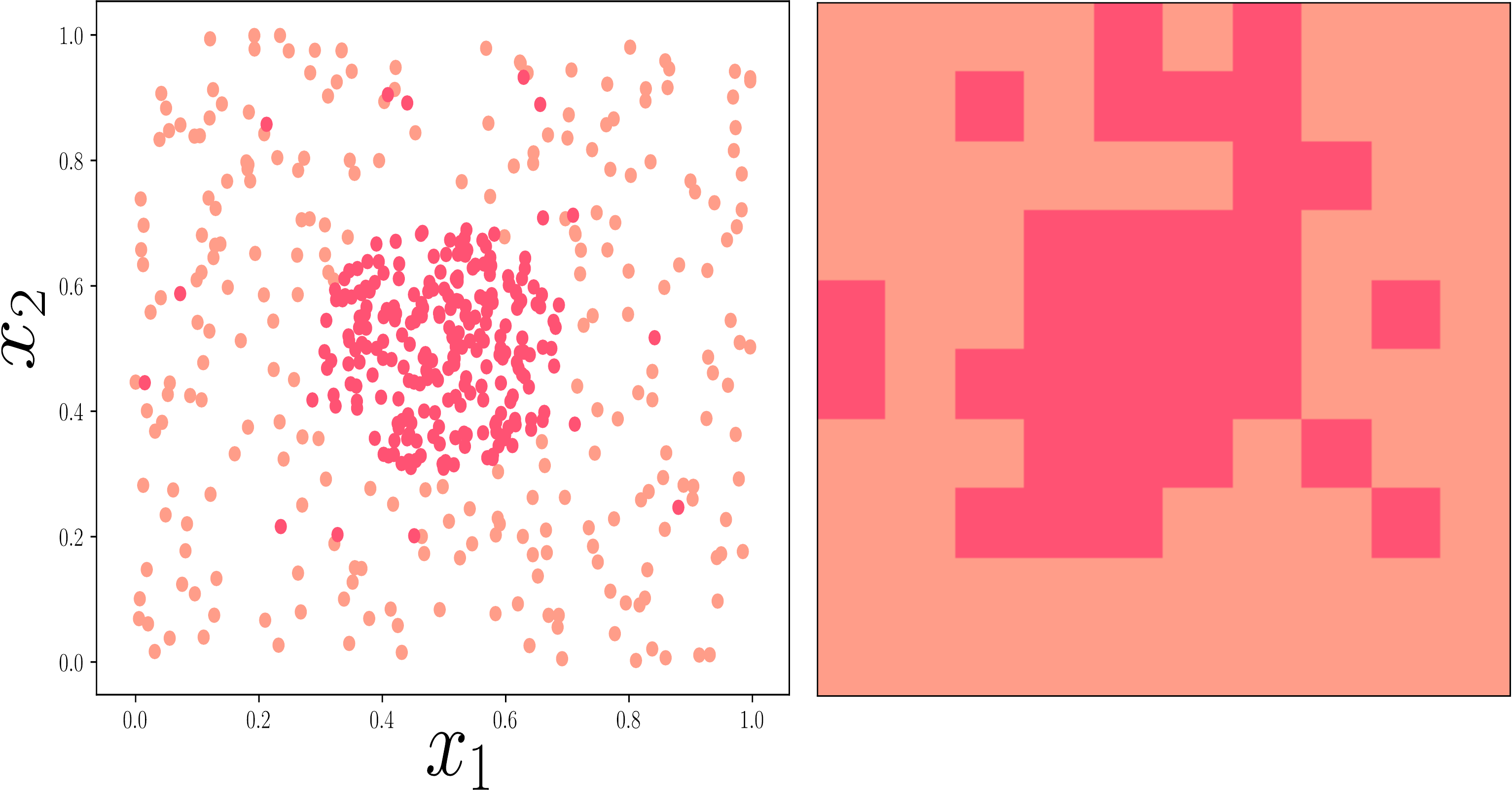}\\
		\small (a) & \small (b)\\
	\end{tabular}
	\caption{Data representation in CA: (a) a dataset with $d=2$ dimensions, $|\mathcal{S}|=\{0,1\}$, and $\mathcal{G}=2$ ``bins'', where $\mathbf{X}_t=(X_t^1,X_t^2)$ falls between $[3,7]$ (min/max $X_t^1$) and $[-3,-3]$ (min/max $X_t^2$); (b) data instances are mapped in the grid of a cellular automaton with $d=2$ and $\mathcal{G}=10$.}
	\label{fig_total_CA}
\end{figure}

Therefore, when adapting a CA for DSM, the above restrictions must be taken into account to yield a CA capable of learning incrementally, and with drift detection and adaptation mechanisms. In order to use CA in DSM, data instances flowing over time must be mapped incrementally to the cells of the grid. Next, we analyze each of the mutually interdependent parts in CA for DSM:
\begin{itemize}[leftmargin=*]
	\item {\verb|Grid:|} In a data mining problem with $n$ features, the standard procedure adopted in the literature consists of assigning one grid dimension to each feature. After that, it is necessary to split each grid dimension by the values of the features, in a way that we obtain the same number of cells per dimension. To achieve that, ``bins'' must be created for every dimension (Figure \ref{fig_total_CA}) by arranging evenly spaced intervals based on the maximum and minimum values of the features. These ``bins'' delimit the boundaries of the cells in the grid.
	\item {\verb|States:|} We have to define a finite number of discrete states $|\mathcal{S}|$, which will correspond to the number of labels (classes) considered in the data mining problem.
	\item {\verb|Local rule:|} In data mining the update rule $f(\cdot)$ can adopt several forms. One of the most accepted variants is a majority vote among neighbors' states (labels). For example, for $d=2$:
	\begin{equation*}\label{eq_majvot}
	S([i,j])=\argmax_{s\in \mathcal{S}}\sum_{[k,l]\in f_{\boxplus}([i,j])}\mathbb{I}(S([k,l])=s),
	\end{equation*}
	where the value of $f_{\boxplus}([i,j])$ will be the coordinates of neighboring cells of $[i,j]$, and $\mathbb{I}(\cdot)$ is an auxiliary function taking value $0$ when its argument is false and $1$ if it is true.
	\item {\verb|Neighborhood:|} a neighborhood and its radius must be specified. Even though a diversity of neighborhood relationships has been proposed in the related literature, the ``von Neumann'' (see Figure \ref{vn_neigh}) or ``Moore'' are arguably the most resorted definitions of neighborhood for CA.
	\item {\verb|Initialization:|} the grid is seeded with the feature values of the instances that belong to the training dataset. In order to decide the state of each cell, we assign the label corresponding to the majority of training data instances with feature values falling within the range covered by the cell. After that, cells of the grid are organized into regions of similar labels (Figure \ref{fig_total_CA}).
	\item {\verb|Generations:|} when the initialization step finishes, some cells may remain unassigned, i.e. not all of them are assigned a state (label). In other words, the training dataset used to prepare the CA for online learning might not be large enough to ``fill'' all cells in the grid. In such a case, it becomes necessary to ``evolve'' the grid several times (generations) until all cells are assigned a state. In this evolving process, each cell calculates its new state by applying the update rule over the cells in its neighborhood. All cells apply the same update rule, being updated synchronously and at the same time. Here lies the most distinctive characteristic of CA: the update rule only inspects its neighboring cells, being the processing entirely local (Figure \ref{vn_neigh}).
\end{itemize}

\section{Proposed Approach: CURIE}\label{proposed}

\begin{figure}[h]
	\centering
	\begin{tabular}{c}
		\includegraphics[width=0.7\linewidth]{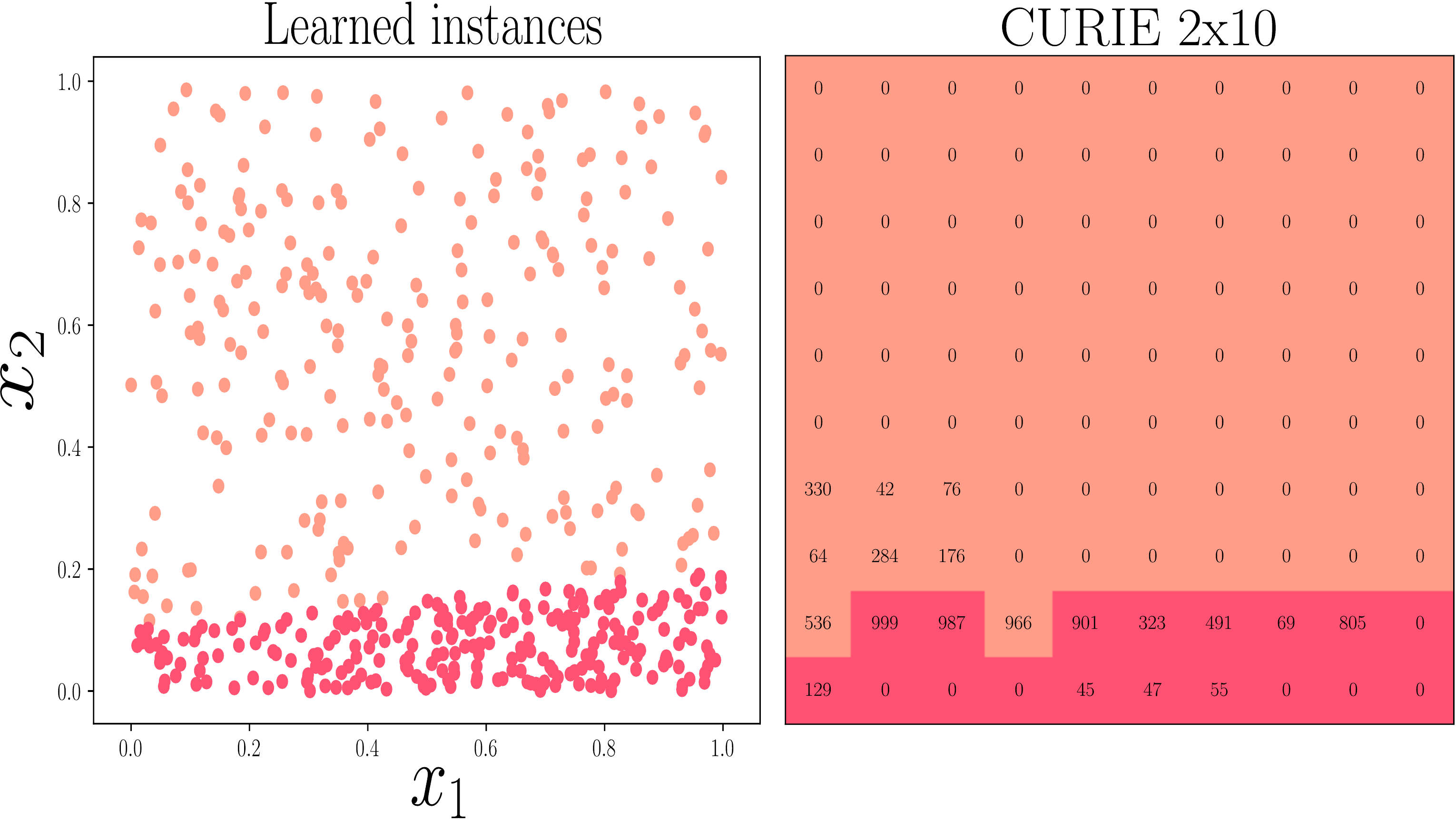}\\ \small (a)\\
		\includegraphics[width=0.7\linewidth]{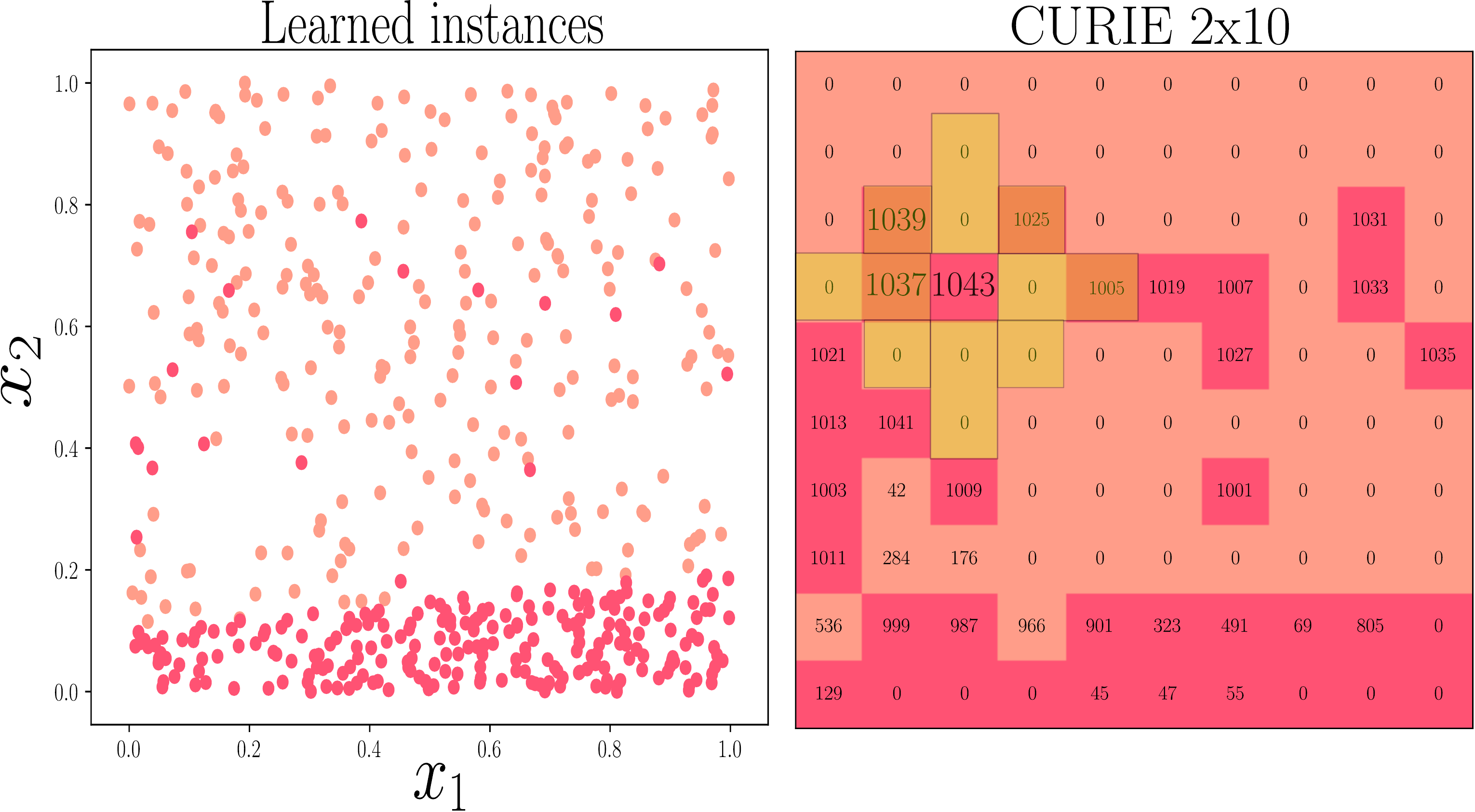}\\\ \small (b)\\
	\end{tabular}
	\caption{The interpretable adaptation mechanism of $\algname$ ($d\times \mathcal{G} = 2 \times 10$) based on the mutations of its neighborhood: (a) before the drift. $\algname$ updates the time instants of each mutant cell, i.e. when the previous state of the cell (before the arriving of the incoming instance) is different from the label of the incoming instance itself; (b) drift occurs. $\algname$ checks the neighborhood of each cell, and when at least $2$ neighboring cells (defined by $n\_muts\_allowed$ parameter) have mutated in the last $10$ time steps (as per the $mutation\_period$ parameter), $\algname$ considers that a drift has occurred. This is what is declared at $t=1043$ with the cell $[2,6]$ and its neighborhood of $r=2$ (Manhattan distance), where $2$ of its neighbors have mutated at time steps $1037$ and $1039$.}
	\label{muts_functioning}
\end{figure}

We delve into the ingredients of $\algname$ to act as drift detector. As shown in Figure \ref{muts_functioning}, its detection mechanism hinges on the evidence that a recent number of mutations in the neighborhood of a cell that has just mutated, may serve as an symptomatic indicator of the occurrence of a drift. 

$\algname$ builds upon this intuition to efficiently identify drifts in data streams by fulfilling the following methodological steps (Algorithm \ref{algorithm_proposed}):
\begin{algorithm}[h]
	\DontPrintSemicolon
	\SetKwInOut{Input}{Input}
	\SetKwInOut{Output}{Output}	
	\Input{Preparatory data instances $[(\textbf{X}_t,y_t)]_{t=0}^{t=P-1}$; training/testing data for the rest of the stream $[(\mathbf{X}_{t},y_t)]_{t=P}^{\infty}$; the grid size $\mathcal{G}$ (``bins'' per dimension); a local update rule \smash{$f_{\circlearrowright}(\cdot)$}; neighborhood function \smash{$f_{\boxplus}(\mathbf{c})$} for cell with coordinates \smash{$\mathbf{c}\in \mathcal{G}=\{1,\ldots,G\}^d$}; radius $r$ for the neighborhood operator; radius $r_{mut}$; maximum number of allowed mutants $n\_muts\_allowed$; time period $mutation\_period$; sliding window $W$ of size $P$.}
	\Output{Trained $\algname$ producing predictions $\widehat{y}_t$ $\forall t\in[P,\infty)$}
	Let $d$ be equal to the number of features in $\mathbf{X}_t$\; 
	Let $|\mathcal{S}|$ be the number of classes (alphabet of $y_t$)\;	
	Set a vector of state hits per cell: $\mathbf{h}_\mathbf{c}=[]$ $\forall \mathbf{c}\in\mathcal{G}$, and a vector of mutations per time step and cell: $\mathbf{h}_\mathbf{m}=[]$ $\forall \mathbf{m}\in\mathcal{G}$\;
	Initialize the limits of the grid: \smash{$[(lim_n^{low},lim_n^{high})]_{n=1}^d$}\;
	Create the grid as per $\mathcal{G}$, $n$ and \smash{$[(lim_n^{low},lim_n^{high})]_{n=1}^d$}\;
	\For(\tcp*[h]{Preparatory process}){$t=0$ to $P-1$}{
		Update limits as per $\mathbf{X}_{t}$, e.g., $lim_n^{low}=\min\{lim_n^{low},x_t^n\}$\;
		Update grid ``bins'' as per $\mathcal{G}$ and \smash{$[(lim_n^{low},lim_n^{high})]_{n=1}^d$}\;
		Select the cell $\mathbf{c}$ in the grid that encloses $\mathbf{X}_{t}$\;
		Append $y_t$ to the vector of state hits $\mathbf{h}_{\mathbf{c}'}=[\mathbf{h}_{\mathbf{c}'},y_t]$\; 	
	}
	Iterate with $r$ and check $|\mathbf{h}_{\mathbf{c}}|$ to ensure one state per cell in $\mathcal{G}$\;	 
	Guarantee at least $|\mathbf{h}_{\mathbf{c}}|=1$ in all cells in $\mathcal{G}$\;
	Iterate with $r$ and recheck $|\mathbf{h}_{\mathbf{c}}|$ to ensure one state per cell\;
	\For(\tcp*[h]{DSM processing}){$t=P$ to $\infty$}{
		Update $W$ with the incoming instance $(\textbf{X}_t,y_t)$\;
		Predict $\widehat{y}_t$ as $S(\mathbf{c})$, with $\mathbf{c}$ denoting the coordinates of the cell enclosing $\mathbf{X}_t$\;
		Update limits as per $\mathbf{X}_{t}$, e.g., $lim_n^{low}=\min\{lim_n^{low},x_t^n\}$\;		
		Update ``bins'' as per $\mathcal{G}$ and \smash{$[(lim_n^{low},lim_n^{high})]_{n=1}^d$}\;
		Save the current cell state: $cur\_st=S(\mathbf{c})$\;
		Update $S(\mathbf{c})=y_t$ (i.e. the verified class of test instance) \;
		\If(\tcp*[h]{A mutation occurs in cell}){$cur\_st \neq y_t$}{
			Append $t$ to the vector of mutations:  $\mathbf{h}_{\mathbf{m}'}=[\mathbf{h}_{\mathbf{m}'},t]$\;
			Calculate \# mutant neighbors $n\_muts$ of the cell, within radius $r_{mut}$ and within time $mutation\_period$\;
			\If(\tcp*[h]{Detection}){$n\_muts >= n\_muts\_allowed$}{
				Initialize $\mathbf{h}_\mathbf{m}, \mathbf{h}_\mathbf{c}$\;
				Initialize grid limits: \smash{$[(lim_n^{low},lim_n^{high})]_{n=1}^d$}\;
				New grid as per $\mathcal{G}$, $n$, \smash{$[(lim_n^{low},lim_n^{high})]_{n=1}^d$}\;	
				Preparatory process ($6-10$) with instances in $W$\;
			}
		}
	}
	\caption{Steps of $\algname$ for drift detection and DSM}\label{algorithm_proposed}
\end{algorithm}

\begin{itemize}[leftmargin=*]
	\item First, the CA inside $\algname$ is created by setting the value of its parameters (detailed as inputs in Algorithm \ref{algorithm_proposed}), and following the characteristics of the given dataset (lines $1$ to $5$).
	\item A reduced number of \emph{preparatory} instances of the data stream  $[(\mathbf{X}_t,y_t)]_{t=0}^{P-1}$) is used to initialize the grid of $\algname$. This grid is seeded with these instances, and then $\algname$ is evolved for several iterations (generations) by applying the local rule until all cells are assigned a state i.e. the labels of the preparatory instances (lines $6$ to $10$). 
	\item When the preparatory process is finished, we must ensure that several preparatory data instances have not seeded the same cell, because each cell must reflect only one single state. To this end, we must assign to each cell the most frequent state by inspecting the labels of all those instances that fell within its boundaries. Then, we must ensure that all cells have an assigned state by applying the local rule iteratively all over the grid. Since this last process can again seed a cell with several instances, we have to address this issue to ensure that the cell only reflects one single state (lines $11$ to $13$).
	\item Next, $\algname$ starts predicting the data instances coming from the stream in a \textit{test-then-train} fashion \cite{gama2014survey} (lines $14$ to $28$). This process consists of first predicting the label of the incoming instance, and next updating the limits of the cells in the grid should any feature value of the processed instance fall beyond the prevailing boundaries of the grid (lines $16$ to $18$). Secondly, the label of the incoming instance is used for training, i.e. for updating the state of the corresponding cell (line $19$).
	\item In line $15$ $\algname$ stores the incoming instance in a sliding window $W$ of size $P$, which is assumed, as in the related literature, to be small enough not to compromise the computational efficiency of the overall approach.
	\item During the \textit{test-then-train} process, $\algname$ checks if a \emph{mutation} of the cell states has occurred (line $21$). If the previous state of the cell (before the arrival of the incoming instance) is different from the label of the incoming instance, a mutation has happened. When there is a mutation, we assign the current time step to the cell in the grid of time steps (line $22$). Then, $\algname$ checks the state of the neighboring cells in a radius $r_{mut}$ (of a von Neumann's neighborhood) in a specific period of time (line $23$). If the number of neighboring mutants exceeds a threshold (line $24$), $\algname$ considers that a drift has occurred.
	\item After drift detection, it is time to adapt $\algname$ to the detected change in the stream distribution. To this end, we reset the grid, the vector of states, and the vector of time steps in which a mutation was present (lines $25$ to $27$). Finally, the preparatory process is carried out by seeding the grid with the instances stored in the sliding window $W$ (line $28$).
\end{itemize}

Finally, after detailing the ingredients of $\algname$ to act as drift detector, we would like to highlight two improvements over \cite{lobo2020lunar} that positively impact on the learning of data distribution:
\begin{itemize}[leftmargin=*]
	\item If the predicted and the true label do not equal each other, the cell state in $\algname$ is always changed to the class of the incoming instance. Otherwise, if the age of the cell state ($T_{age}$) was considered, this could impact on drift detection resulting in more detection delay.
	\item In $\algname$ there is always one state assigned to each cell, thus it is not necessary to check the state of the closest cell among those with assigned state to provide a prediction. The cost of assigning one state to all cells of the grid is insubstantial; it is just carried out at the preparatory process and when drift is detected. And so we achieve a more simple method that does not need to check the surroundings (neighborhood) of the cell when no state is assigned.
\end{itemize}

The source code of $\algname$ is available at \url{https://github.com/TxusLopez/CURIE}.

\section{Experimental Setup}\label{expems}

In order to assess the performance of $\algname$, we have designed several experiments with synthetic datasets configured with both abrupt and gradual drift versions.

Since drift detectors usually base their detection mechanisms on the prediction results of a base learner, both detection and classification are often set to work together. As it has been already mentioned, $\algname$ does not use the prediction of the base learner. Instead, it estimates the occurrence of the drift by looking at the changes that occur in the neighborhood of cells deployed over a grid that represents the distribution fo data. In our experiments we have accordingly combined three well-known base learners (\texttt{HT}, \texttt{NB} and \texttt{KNN}) with five popular drift detectors including our proposed detector (corr. \texttt{DDM}, \texttt{EDDM}, \texttt{PH}, \texttt{ADWIN}, and \texttt{\algname}). They form $15$ different learning-detection schemes following the algorithmic template shown in Algorithm \ref{learn_and_detect}. Such base learners and drift detection methods have been selected due to their wide use by the DSM community, and the availability of their implementations in the scikit-multiflow framework \cite{montiel2018scikit}. For more information, we refer the reader to \cite{gonccalves2014comparative} and \cite{barros2018large}. Please note that the inclusion of \texttt{KNN} is not only based on its widely use, and it has also been considered due to its similarities with CA. While \texttt{KNN} is not strictly local (the neighborhood is not fixed beforehand and an the nearest neighbor of an instance may change), CA has a fixed neighborhood. In CA the local interaction between cells affects the evolution of each cell. We would also like to underline that the size of the sliding window of \texttt{KNN} (\textit{max\_window\_size} parameter in Table \ref{params}) is the same than the number of recent instances that CA uses to be initialized and seeded after a drift is detected.

The computer used in the experiments is based on a $x86$ $64$ architecture with $8$ processors Intel(R) Core(TM) $i7$ at $2.70$GHz, and $32 DDR4$ memory running at $2,133$ MHz. The source code for the experiments is publicly available at this GitHub repository: \url{https://github.com/TxusLopez/CURIE}.

\subsection{Datasets}

In order to assess the performance of a drift detector by measuring the different detection metrics, we need to know beforehand where a real drift occurs. This is only possible with synthetic datasets. The scikit-multiflow framework \cite{montiel2018scikit} allows generating several kinds of synthetic data to simulate the occurrence of drifts. Concretely, we have generated $20$ diverse synthetic datasets ($10$ abrupt and $10$ gradual) by using several stream generators and functions, and with a different number of features and noise. They exhibit $4$ concepts and $3$ drifts at time steps $10,000$, $20,000$, and $30,000$ in the case of abrupt datasets, and at time steps $9,500$, $20,000$, and $30,500$ in the case of gradual ones. In the latter case, the width of the drift is $1,000$ time steps. All generated data streams have $40,000$ instances in total. Next, the details of the datasets:
\begin{itemize}[leftmargin=*]
	\item With {\verb|Sine|} generator: it is ruled by a sequence of classification functions. \texttt{Sine\_A} refers to abrupt cases and \texttt{Sine\_G} to gradual ones. In the case of \texttt{Sine\_F1}, the order of the functions is SINE1 ,reversed SINE1, SINE2, and reversed SINE2. For \texttt{Sine\_F2} the order is namely, reversed SINE2-SINE2-reversed SINE1-SINE1. Therefore, \texttt{Sine} stream generator provides $4$ different datasets: \texttt{Sine\_A\_F1}, \texttt{Sine\_A\_F2}, \texttt{Sine\_G\_F1}, and \texttt{Sine\_G\_F2}. They consist of $2$ numerical features without noise, and a balanced binary class .
	\item With {\verb|Random Tree|} generator: it is ruled by a sequence of tree random state functions. The parameters \textit{max\_tree\_depth}, \textit{min\_leaf\_depth}, and \textit{fraction\_leaves\_per\_level} were set to $6$, $3$, and $0.15$ respectively. \texttt{RT\_A} refers to abrupt cases and \texttt{RT\_G} to gradual ones. In the case of \texttt{RT\_F1}, the order of the functions is $8873$-$9856$-$7896$-$2563$. For \texttt{RT\_F2} the order is reversed $2563$-$7896$-$9856$-$8873$. Therefore, \texttt{Random Tree} stream generator provides $4$ different datasets: \texttt{RT\_A\_F1}, \texttt{RT\_A\_F2}, \texttt{RT\_G\_F1}, and \texttt{RT\_G\_F2}. They consist of $2$ numerical features without noise, and  balanced binary class.		
	\item With {\verb|Mixed|} generator: it is ruled by a sequence of classification functions. \texttt{Mixed\_A} refers to abrupt cases and \texttt{Mixed\_G} to gradual ones. In the case of \texttt{Mixed\_F1}, the order of the functions is $0$-$1$-$0$-$1$. For \texttt{Mixed\_F2} the order is reversed $1$-$0$-$1$-$0$. Therefore, \texttt{Mixed} stream generator provides $4$ different datasets: \texttt{Mixed\_A\_F1}, \texttt{Mixed\_A\_F2}, \texttt{Mixed\_G\_F1}, and \texttt{Mixed\_G\_F2}. They consist of $4$ numerical features without noise, and a balanced binary class.		
	\item With {\verb|Sea|} generator: it is ruled by a sequence of classification functions. \texttt{Sea\_A} refers to abrupt cases and \texttt{Sea\_G} to gradual ones. In the case of \texttt{Sea\_F1}, the order of the functions is $0$-$1$-$2$-$3$. For \texttt{Sea\_F2} the order is reversed $3$-$2$-$1$-$0$. Therefore, \texttt{Sea} stream generator provides $4$ different datasets: \texttt{Sea\_A\_F1}, \texttt{Sea\_A\_F2}, \texttt{Sea\_G\_F1}, and \texttt{Sea\_G\_F2}. They consist of $3$ numerical features, a balanced binary class, and with the probability that noise will happen in the generation of $0.2$ (probability range between $0$ and $1$).			
	\item With {\verb|Stagger|} generator: it is ruled by a sequence of classification functions. \texttt{Stagger\_A} refers to abrupt cases and \texttt{Stagger\_G} to gradual ones. In the case of \texttt{Stagger\_F1}, the order of the functions is $0$-$1$-$2$-$0$. For \texttt{Stagger\_F2} the order is reversed $2$-$1$-$0$-$2$. Therefore, \texttt{Stagger} stream generator provides $4$ different datasets: \texttt{Stagger\_A\_F1}, \texttt{Stagger\_A\_F2}, \texttt{Stagger\_G\_F1}, and \texttt{Stagger\_G\_F2}. They consist of $3$ numerical features without noise, and a balanced binary class.			
\end{itemize}
Finally, as it is explained in section \ref{CA_dat_min}, it is necessary to create ``bins'' by splitting each grid dimension by the values of the features. For \texttt{Sine} and \texttt{RT} datasets we have used $20$ ``bins'' per dimension, while for the rest of datasets we have used $10$ ``bins'. The values have been found experimentally, just knowing that a small grid is not capable of representing the data distribution (e.g. the grid of Figure \ref{vn_neigh}). Here, we would like to warn other researches by underlining that \texttt{\algname} exhibits at this moment a drawback that should be considered. Due to its exponential complexity, we recommend the use of \texttt{\algname} in datasets with a low number of features. This setback can be tackled by carrying out the search over the grid’s cells by parallelizing this process.

The datasets are available at this Harvard Dataverse repository: \url{https://dataverse.harvard.edu/dataset.xhtml?persistentId=doi:10.7910/DVN/5OWRGB}.

\subsection{Methods and Parameters}

As for $\algname$ we have assigned one grid dimension to each feature of the dataset at hand. We also note that we have used a reduced number of instances to warm up the learning-detection schemes, and also $\algname$ (see Algorithms \ref{algorithm_proposed} and \ref{learn_and_detect}). The number of instances for this purpose usually depends on the memory or processing time restrictions. In our experiments we have considered a very low number of them in order to simulate a very restrictive real-time environment (see parameter $P$ in Table \ref{params}). In all of them, $\algname$ has been configured with a \emph{von Neumann's} neighborhood rather than opting for its \emph{Moore} alternative. A \emph{von Neumann's} neighborhood is linear in the number of dimensions of the instance space, and therefore scales well for problems of high dimensionality. In addition, a \emph{Moore's} neighborhood includes more neighbors, thus we would have to potentially apply the local rule over more cells. This would make the process computationally heavier and less suited for a DSM setting in the preparatory process and after the drift occurs. 

The parameter configurations for the drift detectors under consideration are detailed in Table \ref{params}. The number of preparatory instances ($P$) and the sliding window ($W$) of size $P$ are shared between $\algname$ and the base learners. Concretely, their values are $P=50$ and $w=P$. The values for the base learners parameters have been found through a hyper-parameter tuning process (Grid Search) carried out with these preparatory instances. Finally, Algorithm \ref{learn_and_detect} presents the details of the learning and detection scheme followed by the experiments.

\begin{table}[h!]
	\centering
	\caption{Configuration of base learners and detectors.}
	\label{params}
	\begin{tabular}{ccc}
		\hline 
		\textbf{Detector} & \textbf{Parameters} & \textbf{Value} \\ \midrule
		\multirow{3}{*}{\texttt{DDM}} & \textit{min\_num\_instances} & $30$ \\
		& $\alpha$ (\textit{warning\_level}) & $2.0$   \\
		& $\beta$ (\textit{out\_control\_level}) & $300$ \\
		\cmidrule(l){1-3}
		\multirow{3}{*}{\texttt{EDDM}} & \textit{min\_num\_instances} & $30$ \\
		& $\alpha$ (\textit{warning\_level}) & $0.95$    \\
		& $\beta$ (\textit{out\_control\_level}) & $0.9$ \\
		\cmidrule(l){1-3}
		\texttt{ADWIN} & $\delta$ & $0.002$\\
		\cmidrule(l){1-3}
		\multirow{4}{*}{\texttt{PH}} & \textit{min\_instances} & $30$ \\
		& $\delta$ & $0.005$ \\
		& \textit{threshold} & $50$    \\
		& $\alpha$ & $0.9999$    \\ 
		\cmidrule(l){1-3}
		\multirow{7}{*}{\texttt{\algname}} & $f_{\boxplus}(\cdot)$ & von Neumann \\
		& $f_{\circlearrowright}(\cdot)$ & Majority voting \\
		& $r, r_{mut}$ & $2$,$2$    \\
		& $|\mathcal{S}|$ & $\{0,1\}$ \\
		& $d \times \mathcal{G}$ & $n\_features \times n\_bins$ \\
		& $mutation\_period$ & $10$ \\
		& $num\_mutants\_neighbors$ & $2$ \\ 			
		\bottomrule
	\end{tabular}%
\end{table}

\begin{algorithm}[h!]
	\DontPrintSemicolon
	\SetKwInOut{Input}{Input}
	\SetKwInOut{Output}{Output}	
	\Input{Preparatory data instances $[(\textbf{X}_t,y_t)]_{t=0}^{t=P-1}$; training/testing data for the rest of the stream $[(\mathbf{X}_{t},y_t)]_{t=P}^{\infty}$; a sliding window $W$ of size $P$}.
	\Output{Trained base learners producing predictions $\widehat{y}_t$ $\forall t\in[P,\infty)$}	
	Base learner $\in$ [\texttt{HT}, \texttt{NB}, \texttt{KNN}]\;
	Initialize base learners parameters of Table \ref{params}\;
	Detector $\in$ [\texttt{DDM}, \texttt{EDDM}, \texttt{ADWIN},\texttt{PH},\texttt{\algname}]\;	
	Initialize detectors parameters of Table \ref{params}\;
	\For(\tcp*[h]{Preparatory process}){$t=0$ to $P-1$}{
		\If{$detector = \texttt{\algname}$}{
			Train detector with $(\textbf{X}_t,y_t)$\;
		}
		Train base learner with ($\mathbf{X}_{t},y_t$)\; 	
	}
	\For(\tcp*[h]{DSM processing}){$t=P$ to $\infty$}{
		Update $W$ with the incoming instance $(\textbf{X}_t,y_t)$\;
		Predict $\widehat{y}_t$\;	
		Train base learner with $(\textbf{X}_t,y_t)$\;			
		\uIf{$detector = \texttt{\algname}$}{
			Train detector with $(\textbf{X}_t,y_t)$\;
		}
		\Else{
			\uIf{$\widehat{y}_t \neq y_t$}{
				detector.add\_element($0$)\;
			}
			\Else{
				detector.add\_element($1$)\;
			}				
		}					
		\If(\tcp*[h]{Detection}){detector.detected\_change()}{
			Initialize detector\;	
			Preparatory process ($6-7$) with instances in $W$\;
		}
	}
	Compare classification and detection performance metrics\;
	\caption{Learning-detection scheme}
	\label{learn_and_detect}
\end{algorithm}

\subsection{Performance Metrics}

Regarding the classification accuracy, we have adopted the so-called \textit{prequential accuracy} ($pACC$) \cite{dawid1999prequential}, which is widely applied in streaming classification scenarios. This metric evaluates the base learner performance by quantifying the average accuracy obtained by the prediction of each test instance before its learning in an online \textit{test-then-train} fashion. This accuracy metric can be defined as: 
\begin{equation*}
pACC(t)=\left\lbrace
\begin{array}{ll}
pACC_{ex}(t)\mbox{, if $t= t_{ref}$; otherwise} \\
\\
preACC_{ex}(t-1) + \dfrac{pACC_{ex}(t)-pACC_{ex}(t-1)}{t - t_{ref} + 1},
\end{array}\right.
\end{equation*}
where $pACC_{ex}(t)=0$ if the prediction of the test instance at time $t$ before its learning is wrong, and $1$ when it is correct. The reference time $t_{ref}$ fixes the first time step used in the estimation, and allows isolating the computation of the prequential accuracy before and after a drift has occurred.

To know about the resources used by stream learners, we have adopted the measure RAM-Hours proposed in \cite{bifet2010fast}, based on rental cost options of cloud computing services. Here, $1$ RAM-Hour equals $1$ GB of RAM dispensed per hour of processing (GB-Hour). In order to analyze the \textit{concept drift} identifications we have used several detection metrics based on true positives (TP), false positives (FP), true negatives (TN), and false negatives (FN):
\begin{itemize}[leftmargin=*]
	\item {\verb|Precision:|} defined as TP$/$(TP$+$FP), is the proportion of predicted drifts that are real drifts.
	\item {\verb|Recall:|} defined as TP$/$(TP$+$FN), is the proportion of the real drifts that have been correctly detected.	
	\item {\verb|Matthews correlation coefficient|} ($MCC$): it is a correlation coefficient between the current and predicted instances. It returns values in the $[-1,1]$ range. It is defined as:
	\begin{equation*}
	MCC=\dfrac{((TP \cdot TN)-(FP \cdot FN))}{\sqrt{(TP+FP)\cdot(TP+FN)\cdot(TN+FP)\cdot(TN+FN)}}.
	\end{equation*}
\end{itemize}
We have also measured the distance of the drift detection to the real drift occurrence (\textmu$D$). Finally, it is worth mentioning that the drifts detected within $2\%$ and $10\%$ (for abrupt and gradual drifts respectively) of the concept size after the real drift positions were computed as TP.

\subsection{Statistical Tests}

We have statistically compared the detectors in all datasets by carrying out the Friedman non-parametric statistical test as described in \cite{demvsar2006statistical}. This test is the first step to know whether any of the detectors have a performance statistically different (in prequential accuracy, RAM-Hours, \textmu$D$, and $MCC$) from the others. The null hypothesis states that all detectors are statistically equal, and in all cases was rejected. Then it is necessary to use a post-hoc test to discover in what detectors there is a statistical difference (in prequential accuracy, RAM-Hours, \textmu$D$, and $MCC$), and we used the Nemenyi post-hoc test \cite{nemenyi1962distribution} with $95\%$ confidence to compare all the detectors against all the others. The results are graphically presented showing the critical difference (CD) represented by bars and detectors connected by a bar are not statistically different.

\section{Results and Discussion}\label{res}

For the sake of space, we only present the mean results in Table \ref{tab-results}. Detailed results are provided in \url{https://github.com/TxusLopez/CURIE}.

\begin{table}[h!]
	\centering
	\caption{Mean results and mean ranks of the detectors in each metric for all considered datasets. $pACC$ compiles the prequential accuracy results of those base learners (\texttt{HT}, \texttt{KNN}, and \texttt{NB}) which have been hybridized with each detector (\texttt{DDM}, \texttt{EDDM}, \texttt{ADWIN}, \texttt{PH}, and \texttt{\algname}). $RAM-Hours$ provides the costs of each mentioned hybrid, while \textmu D and $MCC$ show the results for the detection metrics. Note that \textmu D equals $1000$ when there are no true positives; otherwise, if \textmu D would equal i.e. $0$, we would favor this metric.}
	\label{tab-results}
	\begin{tabular}{ccccccc}
		\cline{1-7}
		&                  & \texttt{DDM} & \texttt{EDDM} & \texttt{ADWIN} & \texttt{PH} & \texttt{\algname} \\ \cline{1-7} 
		\multirow{2}{*}{\textbf{pACC}} & \textbf{score}   & $0.81$         & $0.79$          & $0.83$           & $0.81$        & $0.83$           \\ \cline{2-7} 
		& \textbf{rank} & $2.72$         & $4.00$          & $2.18$           & $3.24$        & $2.81$           \\ \cline{1-7} 
		\multirow{2}{*}{\textbf{RAM-Hours}}   & \textbf{score} & $0.0005448$ & $0.0007455$ & $0.0005708$ & $0.0004462$ & $0.0009449$ \\ \cline{2-7} 
		& \textbf{rank} & $3.31$         & $2.56$          & $3.00$           & $2.32$        & $3.82$           \\ \cline{1-7} 
		\multirow{2}{*}{\textbf{\textmu D}} & \textbf{score} & $1000.00$   & $698.80$    & $594.59$    & $892.68$    & $465.45$    \\ \cline{2-7} 
		& \textbf{rank} & $3.81$         & $2.85$          & $2.40$           & $3.54$        & $1.90$           \\ \cline{1-7} 
		\multirow{2}{*}{\textbf{MCC}}  & \textbf{score}   & $0.00$         & $0.06$          & $0.26$           & $0.06$        & $0.37$           \\ \cline{2-7} 
		& \textbf{rank} & $3.93$         & $3.22$         & $2.53$           & $3.56$        & $1.76$ \\ \bottomrule
	\end{tabular}
\end{table}

\begin{figure}[h]
	\centering
	\begin{tabular}{cc}
		\includegraphics[width=0.5\linewidth]{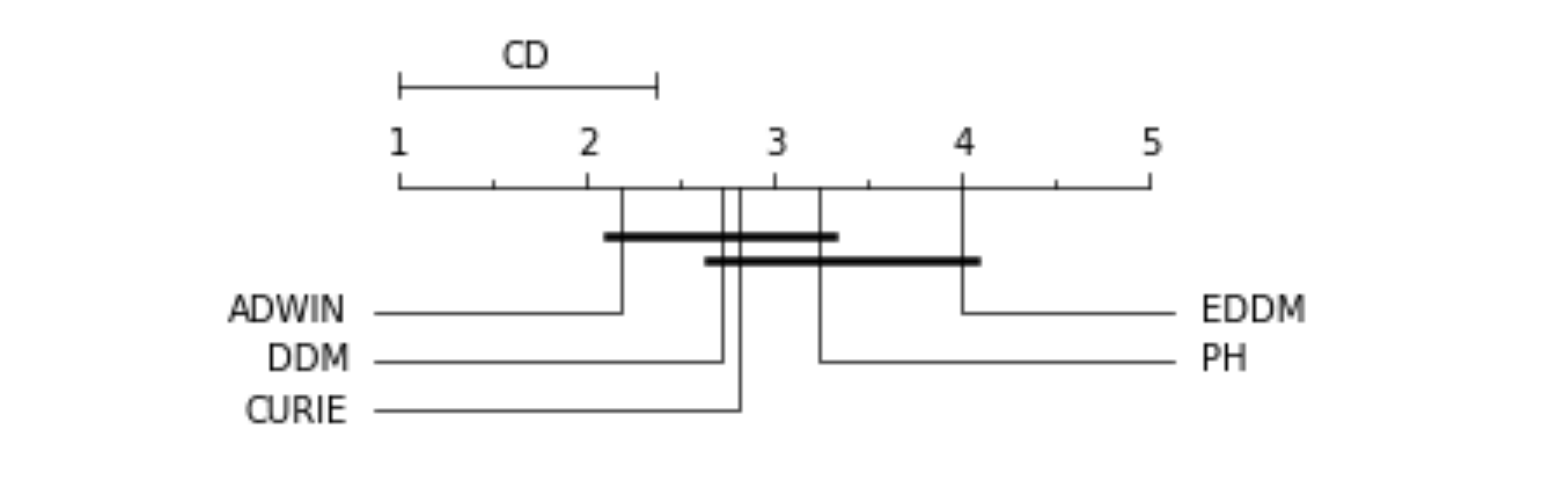} &
		\includegraphics[width=0.5\linewidth]{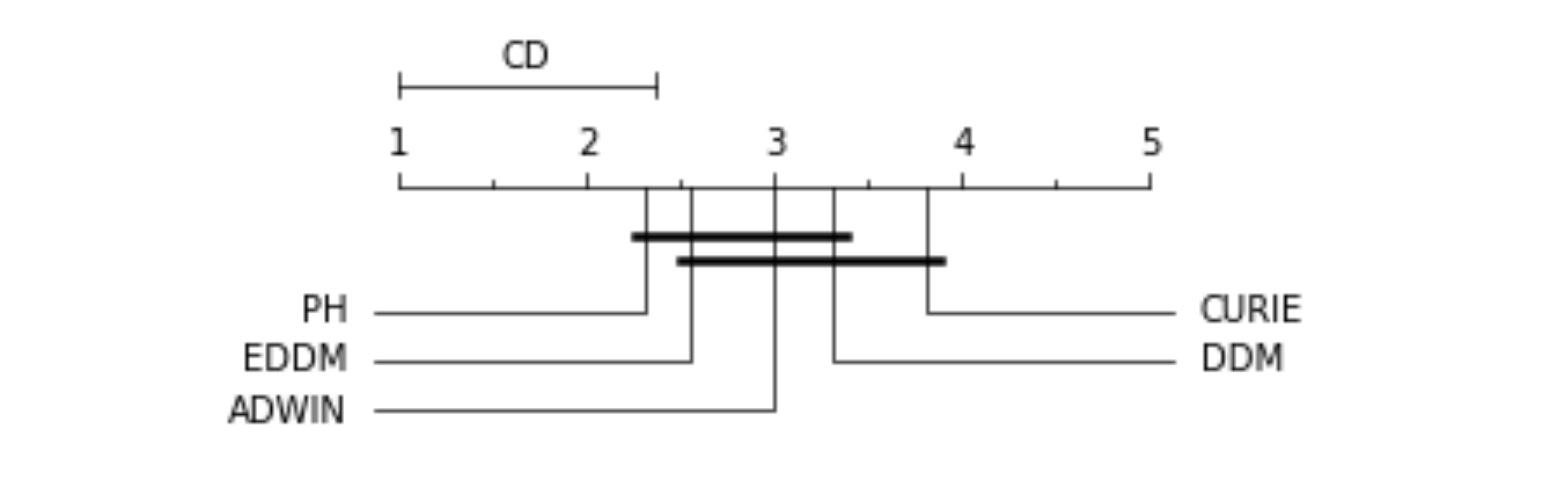}\\
		\small (a) & \small (b)\\
		
		\includegraphics[width=0.5\linewidth]{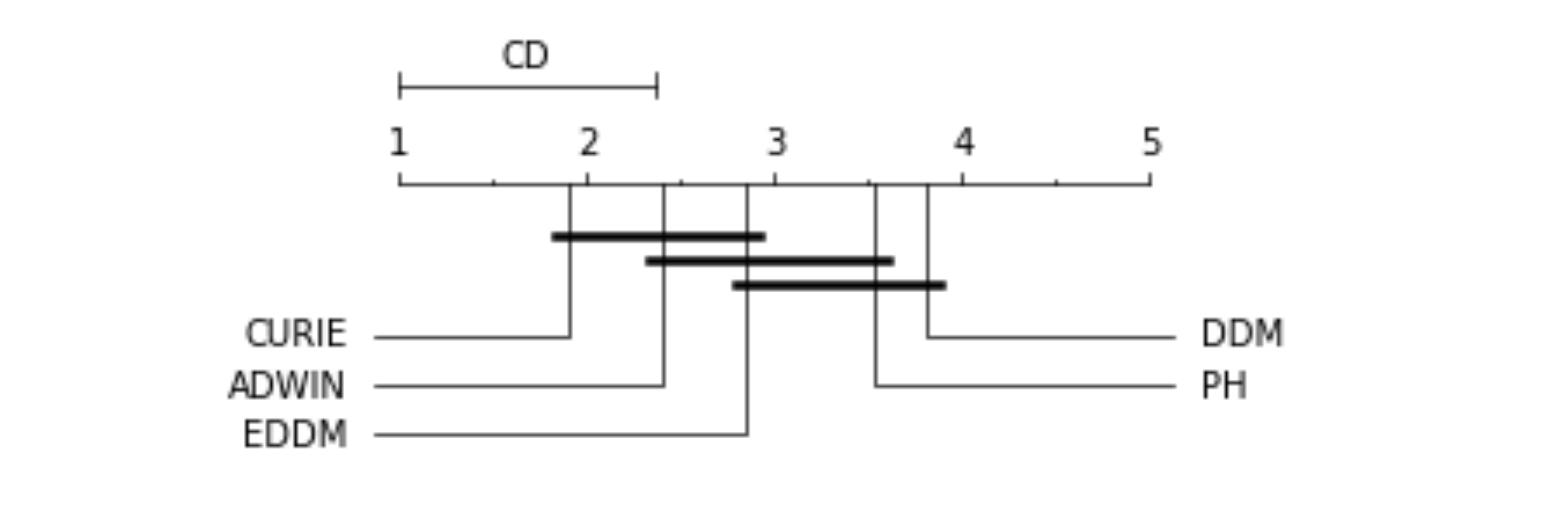} &
		\includegraphics[width=0.5\linewidth]{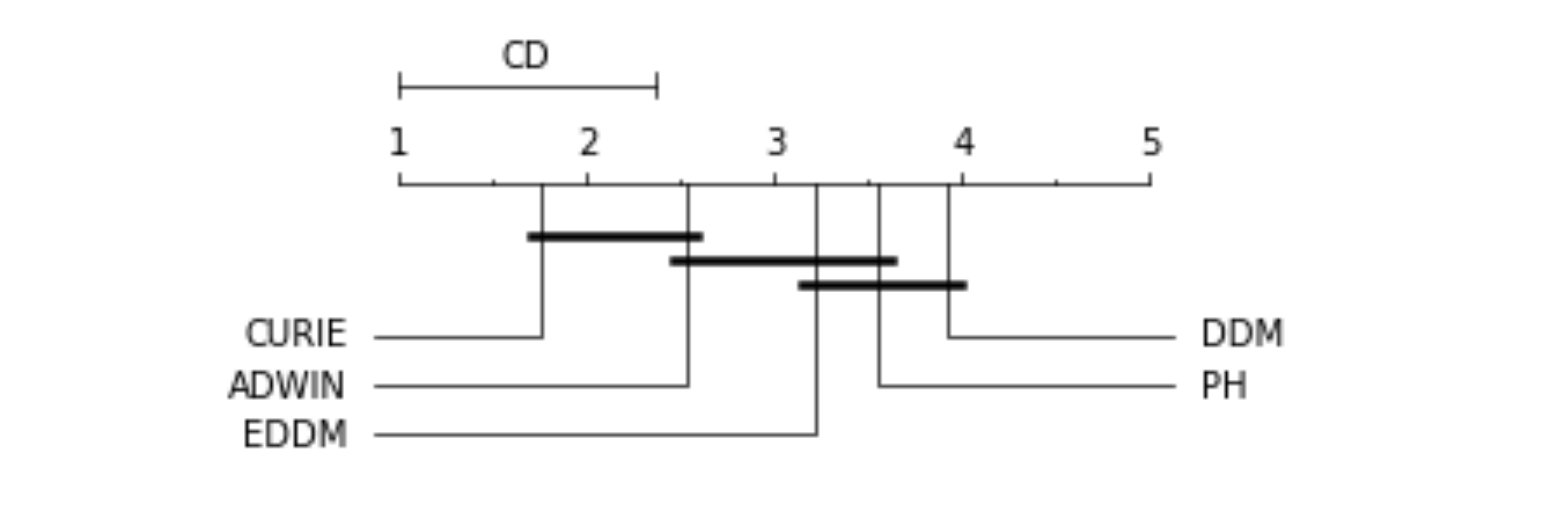} \\
		\small (c) & \small (d)\\
	\end{tabular}
	\caption{Comparison of (a) \textbf{pACC}, (b) \textbf{RAM-Hours}, (c) \textbf{\textmu D}, and (d) \textbf{MCC} results of detectors using the Nemenyi test based on the results of Table \ref{tab-results} with a $95\%$ confidence interval. CD is $1.363887$ for $5$ detectors and $20$ datasets.}
	\label{nemenyi_res}
\end{figure}

Figure \ref{nemenyi_res} presents the evaluation of the concept drift detection methods based on the results of the Table \ref{tab-results}. 

In Table \ref{tab-results} we observe that \texttt{\algname} and \texttt{ADWIN} achieve the best $pACC$ metric with $0.83$. However, \texttt{\algname} is the worst in terms of $RAM-Hours$ with $0.0009449$. Here, in favor of \texttt{\algname}, it is worth mentioning that it is competing with well-established detectors whose code have been optimized and tested by the community in the scikit-multiflow framework \cite{montiel2018scikit}. Probably, future versions of \texttt{\algname} will be more competitive in terms of this metric. Regarding detection metrics, \texttt{\algname} is the best with $465.45$ for \textmu D and $0.37$ for $MCC$.

According to the ranks of Figure \ref{nemenyi_res}, \texttt{ADWIN}, \texttt{DDM} and \texttt{\algname} are the best detectors in terms of $pACC$, yet no statistical differences between them. Regarding the $RAM-Hours$ metric, \texttt{\algname} and \texttt{DDM} are the worst detectors, with no statistical differences between them. However, in what refers to \textmu D and $MCC$, \texttt{\algname}, \texttt{ADWIN}, and \texttt{EDDM} are the best detectors, yet again no statistical differences between them. The conditions of the Nemeneyi test have been very tight ($95\%$ confidence for $5$ detectors in $20$ datasets) and it is difficult to achieve statistical differences. Even so, \texttt{\algname} has shown to be an interpretable drift detector competitive in terms of predictive performance ($pACC$) and detection metrics (\textmu D and $MCC$), without depending on the output (class prediction) of the base learner. Moreover, \texttt{\algname} provides competitive metrics for abrupt and gradual drifts, being this issue very controversial in the drift detection field, as it was shown in \cite{gonccalves2014comparative}.

\section{Conclusion and Outlook}\label{concs}

This work has presented $\algname$, a competitive and interpretable drift detector based on cellular automata. Until now, cellular automata have shown to be suitable solutions for data mining tasks due to their simplicity to model complex systems, being robust to noise and a low-bias method. Besides, they are computationally complete with parallelism capacity, and they already showed competitive classification performances as data mining methods. 

This time, we have focused on their capacity to detect \textit{concept drift}. They have revealed themselves as suitable detectors that achieve competitive detection metrics. They have also allowed base learners to exhibit competitive classification accuracies in a diversity of datasets subject to abrupt and gradual concept drifts. They are suitable candidates to represent data distributions with a few instances, being this ability welcomed in data stream mining tasks where memory and computational resources are often severely constrained. Moreover, $\algname$ can act as an all-in-one approach, in contrast to many other drift detectors which are based on a combination of a base learner method with a detection mechanism.

As future work, we aim to extend the experimental benchmark to more synthetic and real datasets in order to extrapolate the findings and conclusions of this study to different types of drift and more realistic applications. Applying ensemble approaches or even networks of cellular automata are also among our subjects of further study.

\begin{acknowledgements}
This work has received funding support from the ECSEL Joint Undertaking (JU) under grant agreement No 783163 (\textit{iDev40} project). The JU receives support from the European Union's Horizon 2020 research and innovation programme, national grants from Austria, Belgium, Germany, Italy, Spain and Romania, as well as the European Structural and Investment Funds. Authors would like to also thank the ELKARTEK and EMAITEK funding programmes of the Basque Government (Spain).
\end{acknowledgements}

%
\section*{Conflict of interest}
The authors declare that they have no conflict of interest.

\bibliographystyle{plain}      
\bibliography{bliblio}   

%
%

\end{document}